\def\paperID{11948}
\def\confName{ICCV}
\def\confYear{2025}

\def\paperTitle{BREEN: Bridge Data-Efficient Encoder-Free Multimodal Learning with Learnable Queries}

\def\authorBlock{
    $^\spadesuit$Tianle Li,
    $^\varheartsuit$Yongming Rao\thanks{Corresponding authors.},
    $^\varheartsuit$Winston Hu,
    $^\spadesuit$Yu Cheng\footnotemark[1] \\
    $^\spadesuit$The Chinese University of Hong Kong,
    $^\varheartsuit$Tencent Hunyuan Research \\
    {\tt\small tianleli@link.cuhk.edu.hk} \\
    \url{https://github.com/ltl3A87/BREEN}
}

\newif\ifreview 
\newif\ifarxiv \newcommand{\arxiv}{\arxivtrue}
\newif\ifcamera 
\newif\ifrebuttal 

\arxiv % \review OR \arxiv OR \cameraready

\pdfoutput=1
\documentclass[10pt,twocolumn,letterpaper]{article}
\ifreview \usepackage[review]{cvpr} \fi
\ifarxiv \usepackage[pagenumbers]{cvpr} \fi
\ifrebuttal \usepackage[rebuttal]{cvpr} \fi
\ifcamera \usepackage{cvpr} \fi

\usepackage{graphicx}	
\usepackage{amsmath}	
\usepackage{amssymb}	
\usepackage{booktabs}
\usepackage{times}
\usepackage{microtype}
\usepackage{epsfig}
\usepackage{caption}
\usepackage{float}
\usepackage{placeins}
\usepackage{color, colortbl}
\usepackage{stfloats}
\usepackage{enumitem}
\usepackage{tabularx}
\usepackage{xstring}
\usepackage{multirow}
\usepackage{xspace}
\usepackage{url}
\usepackage{subcaption}
\usepackage{xcolor}
\usepackage{fdsymbol}
\usepackage[hang,flushmargin]{footmisc}

\usepackage{pifont}
\newcommand*\colourcheck[1]{%
  \expandafter\newcommand\csname #1check\endcsname{\textcolor{#1}{\ding{52}}}%
}
\colourcheck{green}

\ifcamera \usepackage[accsupp]{axessibility} \fi

% ...

\ifarxiv  \fi

\newcommand{\R}[1]{{%
    \textbf{%
        \ifstrequal{#1}{1}{\textcolor{red}{R#1}}{%
        \ifstrequal{#1}{2}{\textcolor{blue}{R#1}}{%
        \ifstrequal{#1}{3}{\textcolor{magenta}{R#1}}{%
        \ifstrequal{#1}{4}{\textcolor{teal}{R#1}}{%
                           \textcolor{cyan}{R#1}%
        }}}}%
    }%
}}
\usepackage{xr-hyper}

\makeatletter
\newcommand*{\addFileDependency}[1]{
  \typeout{(#1)}
  \@addtofilelist{#1}
  \IfFileExists{#1}{}{\typeout{No file #1.}}
}

\makeatother
\newcommand*{\myexternaldocument}[1]{
    \externaldocument{#1}
    \addFileDependency{#1.tex}
    \addFileDependency{#1.aux}
}

\definecolor{cvprblue}{rgb}{0.21,0.49,0.74}
\usepackage[pagebackref,breaklinks,colorlinks,allcolors=cvprblue]{hyperref}
\usepackage[capitalize]{cleveref}
\crefname{section}{Sec.}{Secs.}
\crefname{table}{Table}{Tables}
\crefname{figure}{Fig.}{Figs.}

\ifarxiv \crefname{appendix}{App.}{Apps.}
\else \crefname{appendix}{Suppl.}{Suppls.} \fi

\unless\ifarxiv \myexternaldocument{_supplementary} \fi

\begin{document}
%% TITLE
\title{\paperTitle}
\author{\authorBlock}
\maketitle

\begin{abstract}
Encoder-free multimodal large language models (MLLMs) eliminate the need for a well-trained vision encoder by directly processing image tokens before the language model. While this approach reduces computational overhead and model complexity, it often requires large amounts of training data to effectively capture the visual knowledge typically encoded by vision models like CLIP. The absence of a vision encoder implies that the model is likely to rely on substantial data to learn the necessary visual-semantic alignments. In this work, we present BREEN, a data-efficient encoder-free multimodal architecture that mitigates this issue. BREEN leverages a learnable query and image experts to achieve comparable performance with significantly less training data. The learnable query, positioned between image and text tokens, is supervised by the output of a pretrained CLIP model to distill visual knowledge, bridging the gap between visual and textual modalities. Additionally, the image expert processes image tokens and learnable queries independently, improving efficiency and reducing interference with the LLM’s textual capabilities. BREEN achieves comparable performance to prior encoder-free state-of-the-art models like Mono-InternVL, using only 13 million text-image pairs in training—about one percent of the data required by existing methods. Our work highlights a promising direction for data-efficient encoder-free multimodal learning, offering an alternative to traditional encoder-based approaches.
\end{abstract}
\section{Introduction}
\label{sec:intro}

% To insert a figure: \input{figs/template}
% Or table: \input{tables/template}

\begin{figure}[t]
\centering
\small
\includegraphics[width=1\linewidth]
{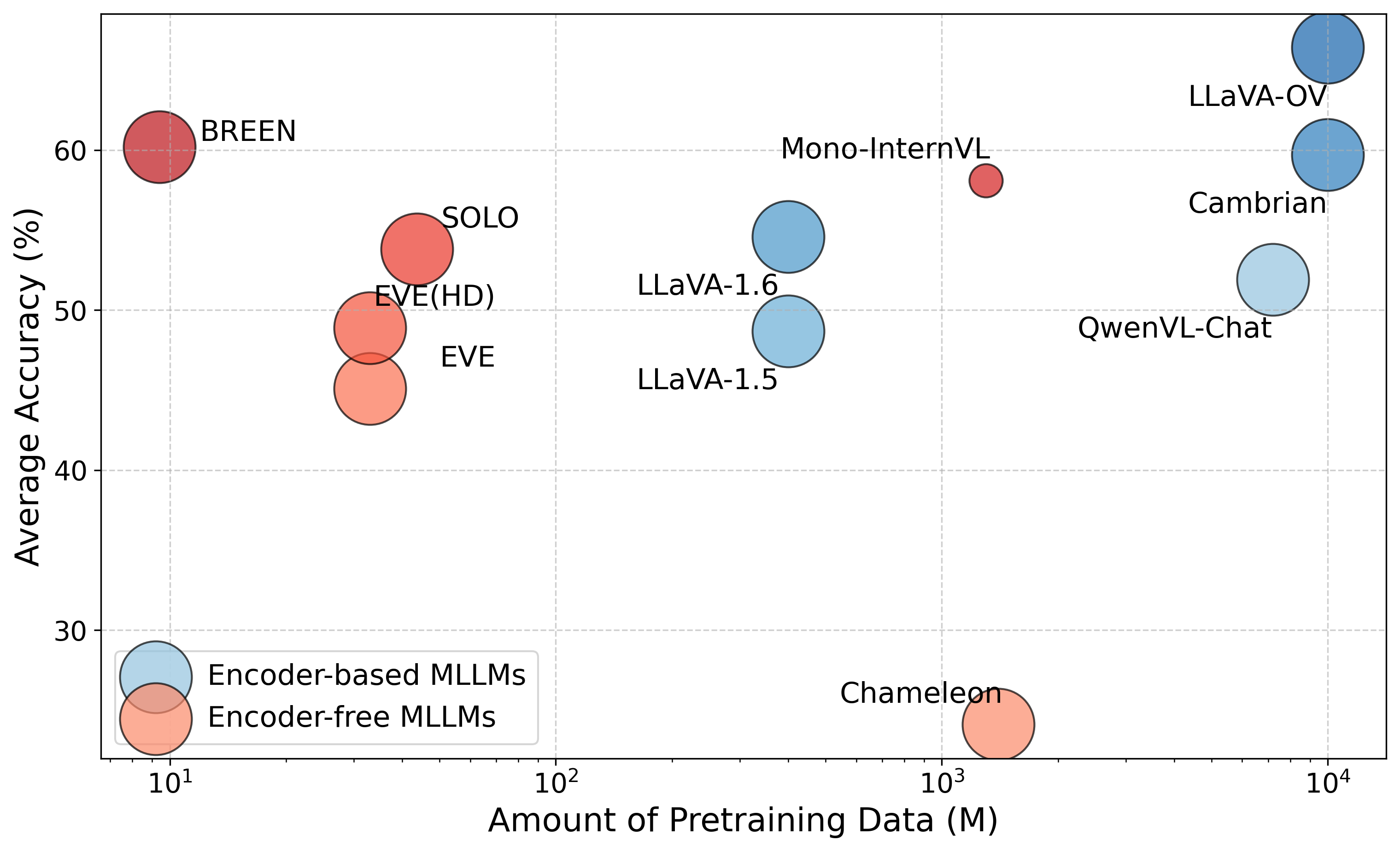}
  \caption{Comparison of encoder-based and encoder-free multimodal large language models (MLLMs) in terms of pretraining data volume, average accuracy on evaluation benchmarks, and model size. The x-axis represents the amount of pretraining data (in millions) on a logarithmic scale, while the y-axis shows the average accuracy (\%). The size of each circle corresponds to the model’s LLM backbone size.}
\label{fig:model-comp}
\end{figure}

\begin{figure*}[t]
    \centering 
    \includegraphics[width=0.98\linewidth,trim= 0 0 0 0,clip]
    {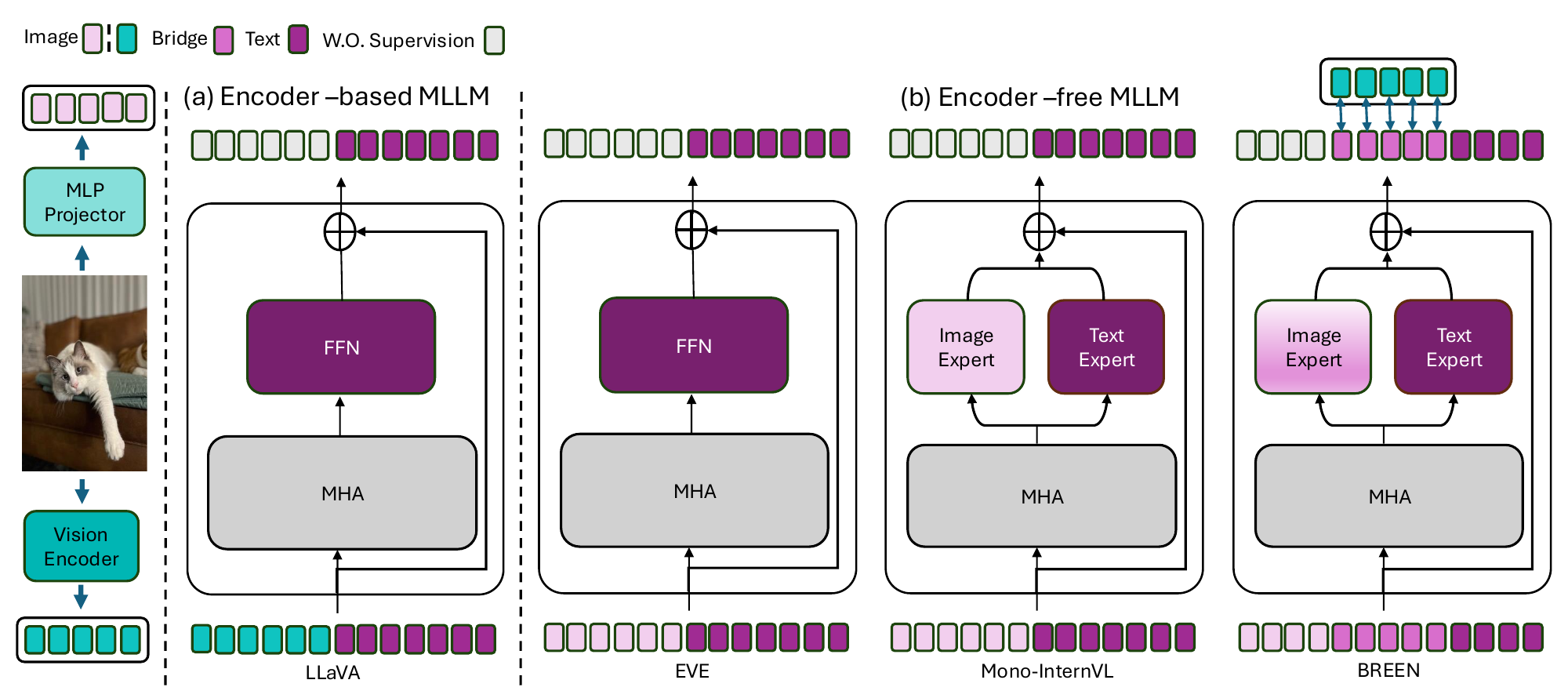} 
    \caption{Comparison among different architectures. For Encoder-based MLLMs, we choose the series of LLaVA~\cite{Model:LLaVA, Model:LLaVA-1.5, Model:LLaVA-NeXT} as the representative. It feeds the images to the well-trained vision encoder, and project the outputs into the input space of the base LLM. For Encoder-free MLLMs, we choose previous works as EVE~\cite{Model:EVE} and Mono-InternVL~\cite{Model:Mono-InternVL} to compare. Instead of encoding the images with a vision encoder, these models use simpler and shallow patch embedding layers to encode the image patch directly.}
    \label{fig:compare}
\end{figure*}

Multimodal large language models (MLLMs) have shown impressive capabilities in understanding and generating text conditioned on visual inputs~\cite{Model:DeepSeek-VL, Model:Gemini, Model:Qwen2-vl, Model:GPT-4, Model:Claude3, Model:InternVL25}. Traditional approaches rely on vision encoders, such as CLIP~\cite{Model:CLIP}, to extract encoded visual representations, which are then fused with textual inputs within a language model as shown in section (a) in Figure~\ref{fig:compare}~\cite{Model:LLaVA, Model:LLaVA-1.5, Model:LLaVA-1.6}. While this paradigm achieves strong performance across various vision-language tasks, it introduces significant computational overhead and complexity, particularly during inference. The reliance on an explicit vision encoder not only necessitates additional processing steps but also constrains the visual capabilities due to information loss, hindering real-time applications~\cite{Model:SOLO}. Consequently, researchers have explored alternative architectures that aim to reduce computational burden and loss of visual information, while maintaining competitive performance~\cite{Model:EVE, Model:Fuyu-8b, Model:Mono-InternVL, Model:SOLO}.

To address these limitations, encoder-free architectures have recently gained traction as a more integral alternative. Notable pioneering works such as EVE~\cite{Model:EVE} and Fuyu~\cite{Model:Fuyu-8b} eliminate the explicit vision encoder and instead directly feed image patches into the LLM after MLP layers or shallow patch embedding layers. This design simplifies the processing pipeline and enhances inference efficiency by treating image and text inputs within a unified framework. However, despite these advantages, encoder-free models often underperform compared to their encoder-based counterparts. This performance gap is largely attributed to the absence of aligned visual features from well-trained encoders, such as CLIP or SigLIP~\cite{Model:CLIP, Model:Siglip}, which inherently capture high-level semantics from image inputs through extensive pretraining. A concurrent work, Mono-InternVL~\cite{Model:Mono-InternVL}, attempts to mitigate this issue by incorporating image experts that process vision tokens separately from text tokens. Combined with large-scale pretraining on billions of text-image pairs, this approach improves alignment and enhances multimodal reasoning from scratch without relying on the extraction of the image features from vision encoder. However, such extensive pretraining is resource-intensive, taking 256 A100 GPUs 16 days to complete the pretraining.

In this work, we introduce \textbf{BREEN}, a \textbf{data-efficient encoder-free multimodal architecture} that effectively distills knowledge from vision encoders while maintaining strong performance with significantly \textbf{fewer training resources}. Our key innovation is a \textbf{learnable query}, positioned between image and text tokens, which serves as a bridge for transferring semantic knowledge from a pretrained CLIP model during the pre-aligning and pretraining stages. Unlike prior encoder-free models that require massive datasets to learn alignment from scratch, BREEN efficiently \textbf{leverages only 13 million text-image pairs in total for pre-training and SFT}, substantially reducing data requirements while achieving superior performance. To further enhance vision interactions, we introduce an \textbf{image expert} that processes both image tokens and learnable queries independently, minimizing interference with the LLM’s core textual reasoning capabilities. As shown in Figure~\ref{fig:model-comp}, these designs enable BREEN to \textbf{achieve state-of-the-art performance among encoder-free models while being significantly more data-efficient}, avoiding the need for extensive multimodal pretraining.

Additionally, we observe that the granularity of the learnable query is crucial for different multimodal tasks. Fine-grained CLIP features are beneficial for spatially aware tasks such as OCR and geometric reasoning~\cite{Datasets:OCRBench, Datasets:OCRVQA}, whereas coarse-grained representations are more suitable for high-level perceptual tasks like color recognition~\cite{Datasets:MME}. To balance this trade-off, we concatenate both fine-grained and coarse-grained learnable queries into a single sequence, allowing the learnable queries to align with representations of different granularity simultaneously. This mechanism ensures strong performance across a diverse range of multimodal tasks. To further validate the effectiveness of learnable queries in multimodal reasoning and understanding, we analyze the model’s attention patterns on image tokens and learnable query tokens. The visualization demonstrates that learnable queries play a crucial role in refining attention to relevant image regions, which may provide a more intuitive explanation on the reason why the inclusion of the learnable queries can enhance the performance.

With a training dataset comprising approximately \textbf{13 million} text-image pairs across the pre-aligning, pretraining, and SFT stages, BREEN surpasses prior encoder-free models such as EVE(HD) and Mono-InternVL, achieving a \textbf{2\% performance improvement} on average across multiple multimodal benchmarks, including MMMU, MME, MMVet, GQA, AI2D, MMB, MMStar, etc.~\cite{Datasets:GQA, Datasets:MM-vet, Datasets:MME, Datasets:MMBench, Datasets:AI2D, Datasets:MMMU, Datasets:MMStar}.

Our contributions can be summarized as follows:
\begin{itemize}

\item We propose a novel encoder-free architecture BREEN that leverages a learnable query to transfer visual knowledge from a pretrained CLIP model, bridging the performance gap between encoder-based and encoder-free models while maintaining inference efficiency.
\item We discover that the combination of fine-grained and coarse-grained learnable queries can enhance multimodal understanding across diverse tasks. Attention visualizations further demonstrate that learnable queries effectively refine vision-language alignment, improving reasoning capabilities.
\item With only 13 million text-image pairs, BREEN achieves superior results compared to prior encoder-free models, demonstrating the effectiveness of our approach in balancing efficiency and performance.
\end{itemize}

Our work establishes a promising direction toward \textbf{data-efficient} multimodal learning, providing a potential alternative to traditional encoder-based MLLMs.
\section{Related Work}
\label{sec:related}

\subsection{Vision-Encoder-Based Multimodal Models}
Recent advances in large language models (LLMs)~\cite{Model:DeepSeekV3, Model:Qwen2.5, Model:Llama3} have fueled the integration of vision and language modalities, resulting in multimodal large language models. Notable commercial models such as GPT-4V~\cite{Model:GPT-4}, Claude 3.5~\cite{Model:Claude3}, and Gemini~\cite{Model:Gemini}, as well as open-source models like LLaVA~\cite{Model:LLaVA, Model:LLaVA-1.5, Model:LLaVA-1.6}, DeepSeek-VL~\cite{Model:DeepSeek-VL, Model:DeepSeek-VL2}, QwenVL~\cite{Model:Qwen-VL, Model:Qwen2-vl}, and InternVL~\cite{Model:InternVL, Model:InternVL25}, typically rely on vision encoder like CLIP~\cite{Model:CLIP} for visual feature extraction. These models integrate visual and textual information by mapping vision features into the input space of LLMs, achieving strong performance across various vision-language tasks. However, these encoder-based models face significant challenges, particularly related to computational inefficiency and the limitations of pre-trained vision encoders. The reliance on pre-trained visual encoders often results in the loss of task-specific visual information, limiting model flexibility and hindering adaptation to new domains~\cite{Model:SOLO}.

\subsection{Encoder-Free Multimodal Models}

The challenges associated with modular MLLMs have driven research toward encoder-free architectures, which aim to eliminate the need for explicit vision encoders while maintaining strong performance in multimodal tasks. Encoder-free models focuses on obtaining continuous visual tokens through a lightweight structure before feeding them into the LLM. For instance, Fuyu-8B~\cite{Model:Fuyu-8b} processes images directly via a simple linear projection, handling high-resolution images effectively without requiring a vision encoder. Similarly, EVE-7B~\cite{Model:EVE} designs a shallow patch embedding layers to deal with image input and enhance image feature learning through visual alignment in hidden layers. SOLO~\cite{Model:SOLO} presents a simple linear projection to transfer images into continuous embeddings similar to Fuyu. Mono-InternVL~\cite{Model:Mono-InternVL} takes a different approach by embedding new visual parameters into a pretrained LLM, integrating visual experts via a multimodal mixture-of-experts~\cite{Algorithm:MoE} structure. These models represent a possible alternative to traditional vision-encoder-based MLLMs, significantly reducing the computational overhead for the encoding of image features while enabling end-to-end multimodal processing. Another series works like Chameleon~\cite{Model:Chameleon}, Show-o~\cite{Model:Show-o}, and EMU3~\cite{Model:EMU3}, on the other hand, leverage pre-trained discrete visual tokenizer~\cite{Tokenizer:VQGAN, Tokenizer:VQVAE} to encode images into tokens and feed into LLMs after concatenation with text tokens.

\subsection{Data Efficiency in Multimodal Learning}

Data efficiency is essential in the development of multimodal models, especially as training large-scale models becomes resource-intensive. Approaches like Bunny~\cite{Model:Bunny} and ALLaVA~\cite{Model:ALLaVA} emphasize the use of curated datasets or synthetic data to reduce the need for large amounts of labeled data, achieving competitive performance with fewer examples. In contrast, Mono-InternVL~\cite{Model:Mono-InternVL} achieves strong performance by leveraging large datasets (e.g., 1.3B examples) to learn image-text alignment in the absence of a pretrained image encoder. Our work addresses this by utilizing a pretrained image encoder like CLIP and a learnable query to efficiently learn image-text representations with less data. While larger datasets could further improve performance, our method offers a more data-efficient alternative while maintaining strong performance across vision-language tasks.

\section{Method}
\label{sec:method}

\begin{figure*}[t]
    \centering 
    \includegraphics[width=0.98\linewidth,trim= 0 0 0 0,clip]
    {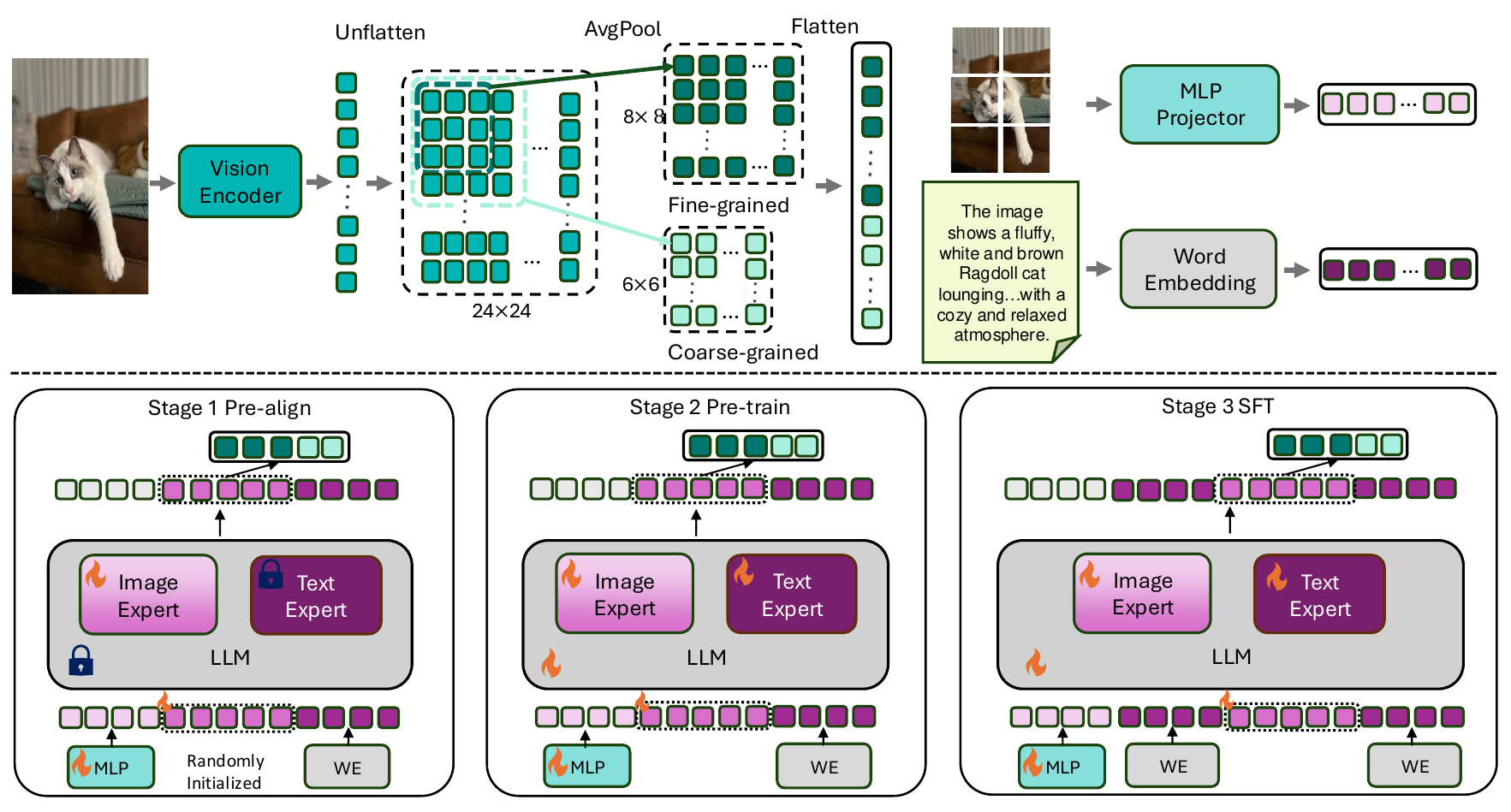} 
    \caption{Overview of the proposed BREEN framework.}
    \label{fig:model}
\end{figure*}

In this section, we describe the architecture design of BREEN and its training methodology as shown in Figure~\ref{fig:model}. BREEN introduces a novel encoder-free multimodal framework that integrates learnable queries in the auto-regressive learning objective and involves an image expert to process both image tokens and learnable query tokens, so that enhancing vision-language alignment. Similarly to previous work~\citep{Model:EVE}, our training pipeline consists of three stages: pre-aligning, pretraining, and supervised fine-tuning (SFT), each contributing to aligning and enhancing the multimodal reasoning capability of the model. 

\subsection{BREEN Architecture}
\label{sec:architecture}

BREEN follows an encoder-free multimodal paradigm, where images are tokenized into patches and directly fed into the LLM. To bridge the modality gap, we introduce a set of learnable query tokens (\(\mathbf{Q}\)) that act as intermediaries between image and text tokens. These queries are supervised by the CLIP-derived visual representations (\(\mathbf{V}_{\text{CLIP}}\)), ensuring effective transfer of visual semantics to the LLM.

\paragraph{Image tokenization and learnable queries} 
As shown in Figure~\ref{fig:model}, images are first divided into patches of size \(P \times P\), which are then projected into embeddings through an MLP layer before being input to the model. Learnable query tokens are inserted between image and text embeddings, facilitating the extraction of meaningful visual knowledge from CLIP. The length of the learnable queries depends on the granularity of the supervision from CLIP \(\mathbf{V}_{\text{CLIP}}\).

To derive structured visual information as supervision for the learnable queries, the input image is resized and padded to CLIP’s required resolution (\(336 \times 336\)), producing a feature representation of size \(24 \times 24\) after being fed to CLIP encoder with a patch size of 14. This feature grid is then unflattened into a two-dimensional structure.

For finer-grained mappings, a patch size and stride of 3 are applied, resulting in a flattened sequence \(\mathbf{V}_{\text{CLIP}}^{\text{fine}}\) of 64 tokens (\(8 \times 8\)) as shown in the upper left part in Figure~\ref{fig:model}. Coarse-grained alignment, using a patch size and stride of 4, produces a sequence \(\mathbf{V}_{\text{CLIP}}^{\text{coarse}}\) of 36 tokens (\(6 \times 6\)). To ensure adaptability across different multimodal tasks, BREEN concatenates randomly initialized fine-grained learnable query \(\mathbf{Q}^{\text{fine}}\) and coarse-grained learnable query \(\mathbf{Q}^{\text{coarse}}\) into a single sequence \(\mathbf{Q}\) of length 100 (64 + 36). These query tokens are positioned after image patches but before text tokens during pretraining, allowing the model to dynamically utilize the appropriate level of visual abstraction through its attention mechanism.

\paragraph{Alignment loss and knowledge transfer} 
To ensure effective knowledge transfer, the output representations of learnable queries (\(\mathbf{Q}_{\text{out}}\)) after LLM processing are aligned with their corresponding CLIP-derived embeddings (\(\mathbf{V}_{\text{CLIP}}\)) using cosine similarity loss:

\begin{equation}
    \mathcal{L}_{\text{align}}^* = 1 - \frac{\mathbf{Q}_{\text{out}}^* \cdot \mathbf{V}_{\text{CLIP}}^*}{\|\mathbf{Q}_{\text{out}}^*\| \|\mathbf{V}_{\text{CLIP}}^*\|}
    \label{eq:align_base_loss}
\end{equation}

where \(\mathbf{Q}_{\text{out}}^*\) denotes the learnable query representations after passing through the LLM and a linear projection, and \(\mathbf{V}_{\text{CLIP}}^*\) represents the corresponding CLIP image embeddings. The symbol \( * \) represents the granularity level (fine or coarse). The overall alignment loss is defined as:

\begin{equation}
    \mathcal{L}_{\text{align}} = \mathcal{L}_{\text{align}}^{\text{coarse}} + \mathcal{L}_{\text{align}}^{\text{fine}}
    \label{eq:align_loss}
\end{equation}

The final training objective combines this loss with the standard autoregressive language modeling loss (\(\mathcal{L}_{\text{LM}}\)) for text tokens, where \(\alpha\) and \(\beta\) is the weights for alignment loss and LM loss:

\begin{equation}
    \mathcal{L} = \alpha\mathcal{L}_{\text{align}} + \beta\mathcal{L}_{\text{LM}}
    \label{eq:all_loss}
\end{equation}

\paragraph{Image expert for modality-specific processing} 
To ensure that image patches and learnable queries are processed effectively without interfering with the LLM’s core text processing, an image expert is introduced into the feed-forward network (FFN) of each transformer layer. This specialized expert enables more refined vision-language interactions while better preserving the LLM’s original text-based reasoning capabilities.

\subsection{Training Strategy}
\label{sec:training}

BREEN undergoes three sequential training stages to optimize its multimodal alignment and understanding.

\paragraph{1. Pre-Aligning Stage} 
In this stage, we initialize the model using a pretrained large language model (LLM) backbone. To stabilize early-stage training, only the learnable query tokens, image linear projector, and image expert weights are updated, while all other parameters remain frozen. The goal is to align the learnable queries with CLIP representation, and simultaneously, force the image linear projector and image expert to align the image features to the caption space. This stage ensures that the new parameters provide a strong starting point for effective image understanding. This alignment helps establish a solid foundation for the subsequent multimodal training, leveraging the well-initialized LLM parameters before fine-tuning the entire model. We collect 9 million of image-text pairs randomly selected from BLIP3-KALE~\citep{awadalla2024blip3kaleknowledgeaugmentedlargescale}, and randomly draw 4 millions out of 9 millions for training in the pre-aligning stage.

\paragraph{2. Pretraining Stage} 
Once the pre-alignment is achieved, all model parameters are unfrozen, and BREEN is trained on larger-scale image-text caption pairs. The model jointly optimizes the alignment loss ($\mathcal{L}_{\text{align}}$) and the language modeling loss ($\mathcal{L}_{\text{LM}}$) to reinforce the connection between visual semantics and textual understanding. In the pretraining stage, we make full use of the 9 millions of image-text pairs as mentioned above from BLIP3-KALE to train the model. 

\paragraph{3. Supervised Fine-Tuning (SFT) Stage} 
In the final stage, BREEN is fine-tuned using multimodal instruction datasets with all the parameters unfrozen. To make the learnable query look ahead at the task context according to the casual mask mechanism, we pre-append the instruction text before the learnable queries and after image patches as shown in the last stage in Figure~\ref{fig:model}, enabling the model to infer the appropriate granularity based on task context as the text instruction and image tokens. In this stage, we gather 4 million of image-text instruction following datasets from open-source public datasets, including Cambrian-1~\citep{Tong2024Cambrian1AF}, LLaVAOneVision~\citep{Li2024LLaVAOneVisionEV}, and Cauldron~\citep{laurençon2024matters} to fine-tune the model.

Through these three stages, BREEN effectively bridges the vision-language gap, achieving strong multimodal reasoning capabilities while preserving the efficiency of encoder-free architectures.

\section{Experiment}
\subsection{Implementation Details}
Our BREEN model is based on Qwen2.5-7B-Instruct, where the weights of image expert are initialized from the feed-forward layer in the LLM. The target of learnable queries are extracted by feeding corresponding images to clip-vit-large-patch14. The learning rates for pre-aligning, pre-training and SFT are $4 \times 10^{-4}$, $4 \times 10^{-5}$, and $4 \times 10^{-5}$ respectively. The batch size for these three stages are 512, 512, and 256 correspondingly. In pre-aligning and pre-training stage, we set \(\alpha = 1\) and \(\beta = 1\) in Equation~\ref{eq:all_loss} for aligning loss and LLM loss. In SFT stage, we set \(\alpha = 0.5\) and \(\beta = 1\) to focus more on text generation. The whole training process takes four nodes of 8$\times$A800 GPUs 16 days to finish.

\subsection{Evaluation Benchmarks}
In this work, we evaluate our model using a diverse set of multimodal benchmarks to assess its performance across various image-text understanding tasks. The evaluation benchmarks include MMMU\cite{Datasets:MMMU}, MMB$^\text{en}$ (MMBench-EN)\cite{Datasets:MMBench}, MMV (MMVet)\cite{Datasets:MM-vet} , MME\cite{Datasets:MME}, SQA$^\text{I}$ (ScienceQA-Img)\cite{Datasets:ScienceQA}, TVQA\cite{Datasets:TextVQA}, HallB (HallusionBench)\cite{Datasets:HallB}, AI2D\cite{Datasets:AI2D}, and MMS (MMStar)~\cite{Datasets:MMStar}. These datasets provide a comprehensive suite of challenges for multimodal models, testing their ability to handle a wide range of reasoning tasks involving both visual and textual inputs. We harness a standard and efficient evaluation toolkit lmms-eval\footnote{\url{https://github.com/EvolvingLMMs-Lab/lmms-eval}.} to conduct the evaluation for BREEN. For the other competitors, some of the results are directly obtained from the Open VLM Leaderboard\footnote{\url{https://huggingface.co/spaces/opencompass/opencompass-llm-leaderboard}}, and the other of them are evaluated with VLMEvalKit~\cite{Eva:vlmevalkit}.

% \begin{figure}[t]
% \centering
% \small
% \includegraphics[width=1\linewidth]
% {figs/rad_plot_intro.png}
%   \caption{Comparison of the performance of BREEN with encoder-based and encoder-free models.}
% \label{fig:model-perform}
% \end{figure}

\subsection{Main Result Analysis}
We compare the performance of BREEN against previous multimodal large language models across various multimodal benchmarks. Table~\ref{tab:multimodal_benchmark} presents the results in terms of accuracy for each task and the overall average scores after normalizing the scale of the tasks. Generally, BREEN achieves an average score of 60.2, outperforming all other encoder-free MLLMs and demonstrating competitive results even against encoder-based models. Although encoder-based models generally achieve higher scores due to explicit vision encoders, BREEN narrows this gap, performing comparably to Cambrian (59.7) while remaining behind LLaVA-OV (66.4), which benefits from significantly larger pretraining data (10B+ image-text pairs). 
% We also visualize the performance as shown in Figure~\ref{fig:model-perform}.

Among encoder-free MLLMs, BREEN consistently achieves the highest accuracy on half of the selected benchmarks and is placed second on the other half. Notably, BREEN improves significantly over Mono-InternVL, with a 9\% increase on MMMU (33.7 → 42.7), a 5.9\% gain on MMBench-EN (65.5 → 71.4), a 7.8\% increase on AI2D (68.6 → 76.4), and a 5.2\% improvement on MMStar (46.0 → 51.2). When placed second, BREEN remains only slightly behind the best encoder-free model most of the time, with a 1.2\% difference on MMVet (40.1 vs. 38.9), 0.2\% on ScienceQA-Img (93.6 vs. 93.4), and 0.8\% on GQA (62.6 vs. 61.8). Although Mono-InternVL has a smaller model size (1.8B vs. 7B), it has undergone pretraining on more than 100 times the data compared to BREEN, demonstrating the latter's remarkable data efficiency while maintaining competitive performance. 

According to the comparison, we can discover that a key advantage of BREEN is its data efficiency. While encoder-based models such as Cambrian and LLaVA-OV rely on 10B+ pretraining samples, BREEN achieves competitive results using only 9M pretraining samples, a fraction of the data required by the encoder-based models.

\begin{figure*}[t]
    \centering 
    \includegraphics[width=0.98\linewidth,trim= 0 0 0 0,clip]
    {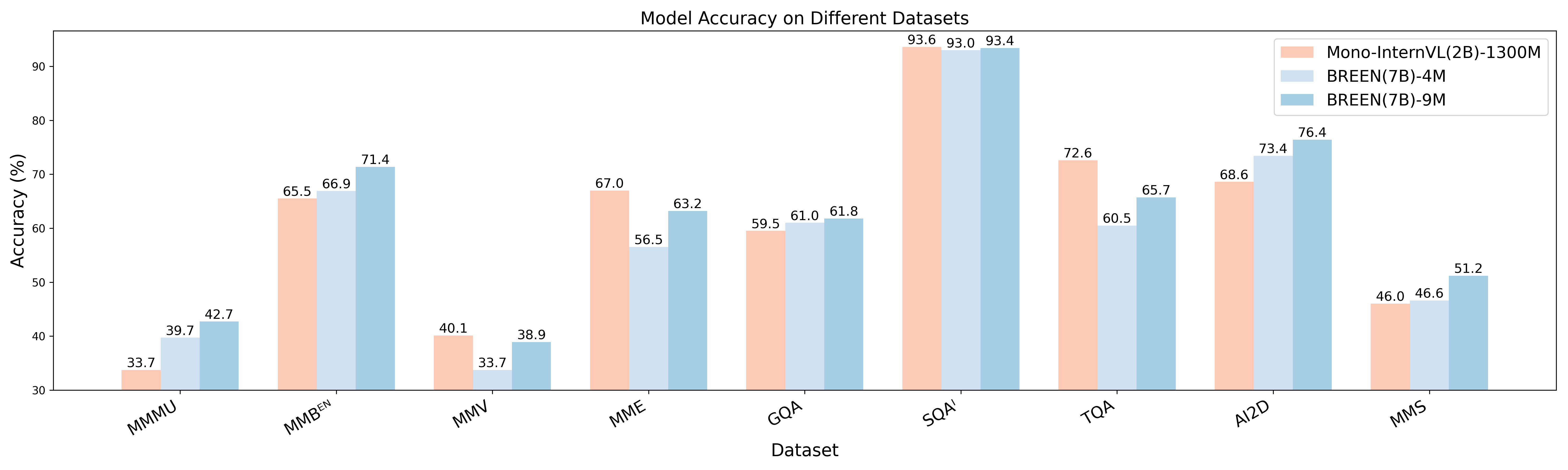} 
    \vspace{-4mm}
    \caption{Comparison on the performance when the amount of pretraining data increases from 4M to 9M}
    \label{fig:data_scaling}
\end{figure*}

We further visualize performance improvements if the amount of data increases from 4M to 9M in the pre-training stage in Figure~\ref{fig:data_scaling}. We can observe that even with merely 4M pretraining data, BREEN can outperform Mono-InternVL on MMMU, MMB, GQA, AI2D and MMStar. In addition, it also shows a great potential that the performance of BREEN can be further boosted if we continue to expand the amount of pretraining data. In a nutshell, despite its much smaller training corpus, BREEN demonstrates strong generalization, particularly excelling in high-level reasoning and multimodal understanding tasks, validating the effectiveness of its learnable query mechanism in distilling multimodal knowledge without requiring an explicit vision encoder.

\begin{table*}[th]
    % \vspace{-2mm}
    % \vspace{-2mm}
    \centering
    \resizebox{\linewidth}{!}{
    \begin{tabular}{l cc| cc cc cccc cc | c}
        \toprule
        Method & \#Param &\#Data 
        & MMMU & MMB$^\text{en}$ 
        & MMV & MME 
        & GQA & SQA$^\text{I}$ & TQA & HallB 
        & AI2D & MMS 
        & Avg\\
        \midrule
        \rowcolor{gray!17}
        \multicolumn{3}{l|}{\emph{Encoder-based MLLMs:}} &\multicolumn{10}{l|}{}&\multicolumn{1}{l}{}\\
        LLaVA-1.5
        & 7B &0.4B+ / 665K 
        & 35.3 & 64.3 
        & 30.5 &1859 
        & 62.0 & 66.8 & 46.1 & 27.6 
        & 54.8 & 33.1
        & 48.7
        \\
        QwenVL-Chat
        & 7B &7.2B / 50M 
        & 35.9 & 60.6 
        & -- &1848 
        & 57.5 & 68.2 & 61.5 & \textbf{36.8}
        & 45.9 & 34.5
        & 51.9
        \\
        LLaVA-1.6
        &7B &0.4B+ / 760K 
        & 35.1 & 67.4 
        & \underline{43.9} &1842 
        & \underline{64.2} & 70.2 & 64.9 & 29.1 
        & 66.6 & 38.4
        & 54.6
        \\
        Cambrian 
        &7B &10B+ / 7M 
        &\underline{42.7} &\underline{75.9} 
        &48.0  &-- 
        &\textbf{64.6} &80.4 &\textbf{71.7} &30.6
        &\underline{73.0} &50.7
        & \underline{59.7}
        \\
        LLaVA-OV 
        &7B &10B+ / 3.2M 
        &\textbf{47.3} &\textbf{81.7}  
        &\textbf{58.8} &\textbf{1998} 
        &-- &\textbf{96.6} &-- & \underline{31.6}
        &\textbf{81.6} &\textbf{61.9}
        &\textbf{66.4}
        \\
        \midrule
        \rowcolor{gray!17}
        \multicolumn{3}{l|}{\emph{Encoder-free MLLMs:}}  &\multicolumn{10}{l|}{}&\multicolumn{1}{l}{}\\
        Chameleon
        &7B &1.4B+ / 1.8M 
        &25.4 &31.1   
        & 8.3 &170 
        & --  & 47.2 &4.8 & 17.1
        &46.0 & 31.1
        & 24.1
        \\
        Fuyu 
        & 8B &-- / -- 
        & 27.9  & 10.7 
        & 21.4 &-- 
        & --   & -- & -- & --  
        & 64.5 & 34.4
        & 31.8
        \\
        EVE
        &7B &33M / 665K 
        & 32.3 & 49.5
        & 25.6 & 1483 
        & 60.8 &63.0 & 51.9 & 21.1 
        & 48.5 & --
        & 45.1
        \\
        EVE(HD)
        &7B &33M / 1.8M 
        & 32.6 & 52.3
        & 25.7 & 1628 
        & \textbf{62.6} &64.9 & 56.8 & 26.4  
        & 61.0 & --
        & 48.9
        \\
        SOLO
        & 8B &43.7M / 2M 
        & --  &  --
        & -- & 1260 
        & --   &73.3   & -- & --  
        & 61.4 & 35.5
        & 53.8
        \\
        Emu3
        & 8B &-- / -- 
        & 31.6 & 58.5
        & 37.2 & 1611
        & 60.3 & 89.2 & 64.7 & 31.7 
        & \underline{70.0} & \underline{46.6} 
        & 54.7
        \\
        Mono-InternVL
        & 1.8B &1.3B / 7M 
        & \underline{33.7}  & \underline{65.5}
        & \textbf{40.1} & \textbf{1875} 
        & 59.5 &\textbf{93.6}  &\textbf{72.6}  & \underline{34.8}  
        & 68.6 & 46.0
        & \underline{58.1}
        \\
        % \textbf{EVEv2.0}
        % & 7B &92M / 7.3M 
        % &\textbf{39.3} &\textbf{66.3} 
        % &\textbf{71.4} &\textbf{45.0} & \underline{1709} & \textbf{87.6}
        % &\textbf{62.9} &\textbf{96.2} &\underline{71.1} &\textbf{73.9}
        % &\textbf{74.8} &\textbf{62.4} &\underline{702}
        % \\
        % \textbf{BREEN}
        % & 7B &4M / 4M 
        % & 39.1 & 66.9 
        % & 33.7 & 1583 
        % & 61 & 93.0 &60.5 &-
        % &73.4 &46.6
        % & --
        % \\
        \textbf{BREEN}
        & 7B &9M / 4M 
        &\textbf{42.7} &\textbf{71.4} 
        &\underline{38.9} & \underline{1770}
        &\underline{61.8} &\underline{93.4} &\underline{65.7} &\textbf{37.0}
        &\textbf{76.4} & \textbf{51.2}
        & \textbf{60.2}
        \\
        \bottomrule
        \\
        \end{tabular}
    }
    \vspace{-2mm}
    \caption{Comparison with previous multi-modal large language models on various multi-modal benchmarks. 
    \#Param denotes the number of parameters of the base LLM models the MLLMs deploy;
    \#Data represents the number of text-images pairs for pre-training / fine-tuning respectively;
    The best results is marked in bold, and the second one is marked with underline.
    }
    \label{tab:multimodal_benchmark}
    % \vspace{-2mm}
\end{table*}

\subsection{Ablation Studies}

\begin{figure*}[t]
    \centering 
    \includegraphics[width=0.98\linewidth,trim= 0 0 0 0,clip]
    {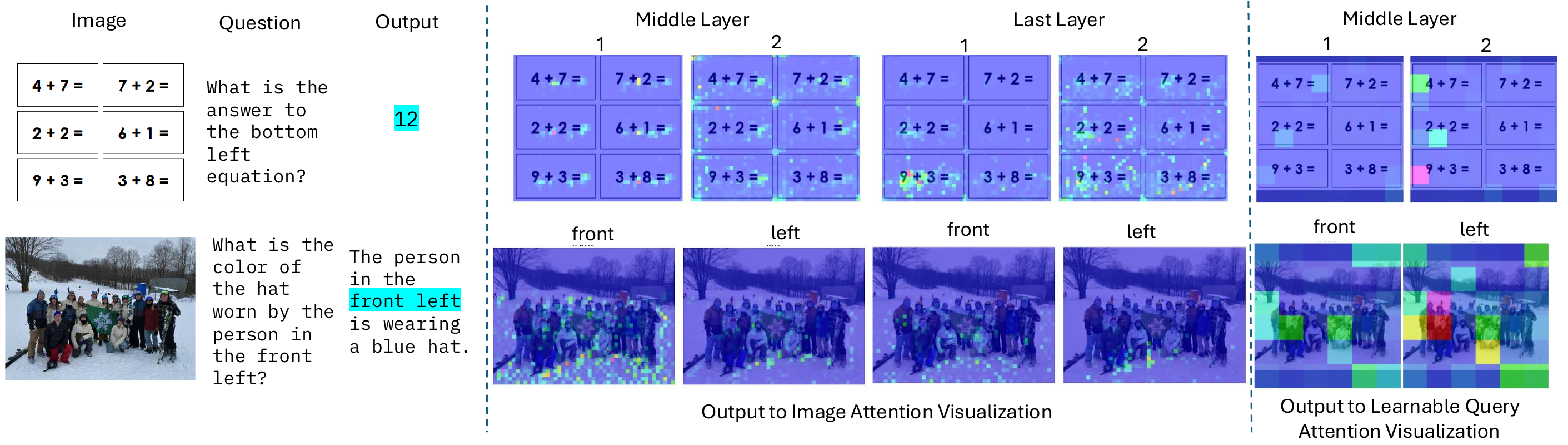} 
    \caption{Visualization of attention scores from output tokens to image tokens and learnable query tokens.}
    \label{fig:atten_viz}
\end{figure*}

\begin{table*}[ht]
\centering
% \footnotesize
\small
\begin{tabular}{ccccccccc}
\toprule
\textbf{Image Expert} & \textbf{LQ(Str2)} & \textbf{LQ(Str3)} & \textbf{LQ(Str4)} & \textbf{SQA$^\text{I}$} & \textbf{TQA} & \textbf{GQA} & \textbf{MME} \\ 
\midrule
&  &  &  & 55.3 & 35.8 & 48.9 & 1243.2 \\
\greencheck &  &  &  & 62.4 & 40.0 & 50.8 & 1260.7 \\
\greencheck & \greencheck &  &  & 62.2 & 41.3 & 50.9 & 1284.3 \\
\greencheck &  & \greencheck &  & 63.3 & 41.1 & 51.7 & 1314.0 \\
\greencheck &  &  & \greencheck  & 62.1 & 41.0 & 51.4 & 1284.9 \\
% \greencheck &  & \greencheck & \greencheck  & 63.5 & 41.1 & 51.6 & 1327.9 \\

\midrule
\greencheck &  & \greencheck & \greencheck & \textbf{63.7} & \textbf{41.4} & \textbf{51.7} & \textbf{1333.4} \\
\bottomrule
 \end{tabular}
    \caption{Ablation study of BREEN. \textbf{LQ(Str2)} means the target of learnable query tokens is extracted from average-pooling of image CLIP output with stride size 2, similar with \textbf{LQ(Str3)} and \textbf{LQ(Str4)}.}
    \label{table:ablation-main}
    \vspace{-4mm}
\end{table*}

\begin{table*}[ht]
\centering
% \footnotesize
\small
\begin{tabular}{cccccccc}
\toprule
\textbf{Stride/Patch Size} & \textbf{Rec} & \textbf{OCR} & \textbf{Know} & \textbf{Gen} & \textbf{Spat} & \textbf{Math} & \textbf{Total} \\ 
\midrule
2 ($12 \times 12$) & 20.9 & \underline{7.8} & 8.6 & 10.6 & 19.1 & 3.8 & 16.2 \\
3 ($8 \times 8$) & 23 & \underline{7.8} & 11.1 & \textbf{13.1} & 18.9 & \textbf{7.7} & 18.1 \\
4 ($6 \times 6$) & \textbf{24.3} & 6.1 & \underline{11.8} & \textbf{13.1} & 18.3 & \textbf{7.7} & \underline{18.6} \\
% 2 \& 4 ($12 \times 12 + 6 \times 6$) & 21.9 & 6.5 & 9.4 & 10 & 16.1 & \textbf{7.7} & 17.4 \\
3 \& 4 ($8 \times 8 + 6 \times 6$) & \underline{23.1} & \textbf{8.6} & \textbf{12} & \underline{12} & \textbf{19.3} & \textbf{7.7} & \textbf{18.7} \\
\bottomrule
 \end{tabular}
    \caption{Ablations of the influence of the stride sizes for learnable queries for different types of tasks in MMVet. The best score is highlighted by bold text and the second to best score is highlighted with the underline.}
    \label{table:ablation-stride}
\end{table*}

To evaluate the contribution of various components in the \textbf{BREEN} architecture, we conduct an ablation study as shown in Table 2. The study focuses on the impact of different architectural decisions, such as the inclusion of the image expert, the use of learnable queries supervised by CLIP outputs at different granularities, and the concatenation of multi-grained learnable queries to balance the performance across various types of tasks. Due to the limited computational resources, we conduct all the ablation studies with the base model Qwen2.5-1.5B-Instruct\footnote{\url{https://huggingface.co/Qwen/Qwen2.5-1.5B-Instruct}}, and leverage a subset of training data, specifically, 4M/4M/1M for the three stages respectively to train the model. The performance metrics are reported across several multimodal benchmarks, including SQA$^\text{I}$, TQA, GQA, and MME.

\paragraph{Impact of Image Expert}
The first and the second row in Table~\ref{table:ablation-main} demonstrates that the inclusion of the image expert significantly boosts the model's performance across all datasets. The image expert, which processes image tokens independently, helps the model better handle the visual modality and aligns it more effectively with textual tokens, leading to notable improvements. For instance, the performance on the \textbf{SQA$^\text{I}$} task jumps from 55.3 to 62.4 with the image expert, while performance on \textbf{TQA} increases from 35.8 to 40.0.

\paragraph{Effect of Learnable Queries with Different Granularity}
When we introduce \textbf{learnable queries} (LQ) based on different granularity of CLIP outputs, we observe a further performance boost. Specifically, in the third, fourth, and fifth rows of Table~\ref{table:ablation-main}, we use learnable queries to align with the representations extracted from CLIP output with different stride sizes (LQ(Str2), LQ(Str3) and LQ(Str4)). The performance of the model on SQA$^\text{I}$ increases from 62.4 to 63.3 with LQ(Str3), and is maintained similar when adding LQ(Str2) or LQ(Str4). For all the other three datasets, we can observe a consistent improvement after integrating learnable queries with different granularity. It is worth noting that not a single granularity of learnable queries can achieve the best performance among all the datasets.

Next, we concatenate the learnable queries from two of the granularity (LQ(Str3) and LQ(Str4)) into a single sequence, as shown in the sixth row of Table~\ref{table:ablation-main}. The performance on each evaluation dataset either maintains as the higher one or surpasses the individual results from stride 3 and stride 4. For example, the performance on \textbf{SQA$^\text{I}$} increases to 63.7 when concatenating the queries, and the accuracy on \textbf{MME} improves to 1333.4. This demonstrates that combining different levels of granularity of visual knowledge with learnable queries allows the model to benefit from the complementary strengths of fine-grained and coarse-grained representations, leading to a more robust multimodal model. Other than directly concatenation, we also explore other ways of multi-grain alignment and ablate the order of granularity, which is demonstrated in Appendix~\ref{sec:appendix_comb}. While BREEN presents a significant step toward data-efficient encoder-free MLLMs, several limitations remain. We discuss the potential limitations and future plans in Appendix~\ref{sec:limiplan}.

% \paragraph{Incorporation of the Binary Mask}
% Finally, in the last configuration (the last row), we introduce a binary mask to selectively attend to different granularity of the visual semantics, depending on the previous input (both image patches and task instruction text). This configuration results in performance improvements across all datasets. For example, accuracy on TQA increases from 40.1 to 41.4, and on MME, it improves from 1327.9 to 1333.4. Overall, the ablation study confirms that the image expert significantly improves performance, and the combination of multiple granularity of learnable queries, along with the use of instruction text ahead and binary masking, provides incremental performance gains. It validates the effectiveness of the BREEN architecture in balancing data efficiency, performance, and task-specific adaptability.

\paragraph{Effect of Granularity for Different Types of Tasks}
To evaluate how different levels of query granularity affect performance across various tasks, we examine the results on the MMVet dataset~\cite{Datasets:MM-vet}, as shown in Table~\ref{table:ablation-stride}. The performance trends of the model vary significantly among tasks with different levels of granularity for the learnable queries. 

For tasks requiring recognition capabilities, such as object identification, we observe that coarser granularity (with a larger stride) leads to better performance. On the other hand, for tasks that demand finer details, such as Optical Character Recognition (OCR) or spatial relationship understanding, finer granularity (with stride 2) achieves superior results. Moreover, the trend of performance changes with granularity is either monotonically increasing or decreasing across the three levels, providing further validation of the observed behavior. When combining granularity with stride values of 3 and 4, we notice a complementary enhancement in performance for tasks like OCR, knowledge retrieval, and spatial reasoning, leading to an overall improvement in performance across the dataset. We also provide the results on different sub-tasks for MME dataset in the Appendix~\ref{sec:appendix_gran}, where similar conclusion can be drawn with MMVet.

\subsection{Visualization and Analysis}

To further investigate the role of learnable queries in multimodal understanding tasks, we visualize the attention scores for both image tokens and learnable query tokens, as shown in Figure~\ref{fig:atten_viz}. The two examples are taken from the MMVet evaluation set and are correctly answered by BREEN. Since BREEN’s base model consists of 28 layers, we select the 14th layer as the middle layer for analysis.

For the visualization of learnable query tokens, we compute the attention score for each query token at the selected layer. We then unflatten the attention score sequence and interpolate each value according to the stride size to map it back to the original input image fed into CLIP. In the first row of Figure~\ref{fig:atten_viz}, the attention scores from the output token to the image in the middle layer highlight the positions of individual equations. In the final layer, the attention weights shift focus more strongly toward the target equation in the bottom left. Similarly, in the visualization of output-to-learnable-query attention, the highest attention score corresponds to the beginning of the target equation (represented by the pink patch below output token ``2'').

In the second row, where the model is tasked with identifying the color of a hat worn by a specific person, the output-to-image attention reveals that the ``front'' and ``left'' tokens do not precisely attend to the desired person in the front left; instead, they broadly distribute attention across those regions. In contrast, the attention to the learnable query accurately pinpoints the target person in the middle layer (indicated by the red patch below the output token ``left''). The attention visualization for more layers of learnable query tokens can be found in Appendix~\ref{sec:viz_attn}. These visualizations and observations provide insights into how the inclusion of learnable queries enhances multimodal understanding tasks.

\section{Conclusion}
\label{sec:conclusion}
In this work, we introduce BREEN, a data-efficient encoder-free multimodal architecture that leverages a learnable query to distill knowledge from a pretrained image encoder, mitigating the limitations of prior encoder-free models. Unlike traditional vision-encoder-based multimodal models, which suffer from computational inefficiencies and rigid visual representations, our approach eliminates the need for a vision encoder while preserving strong visual-text alignment. Compared to existing encoder-free models like Mono-InternVL, which rely on large-scale datasets to learn image-text alignment from scratch, BREEN achieves competitive performance with significantly less data by efficiently transferring semantic information from well-trained image encoders. Experimental results show that BREEN surpasses previous encoder-free models while requiring fewer training resources, highlighting the effectiveness of leveraging pretrained vision models in an encoder-free setting. Future work includes scaling BREEN with larger datasets to explore its full potential, further refining the learnable query for improved representation learning, and extending the model to more complex multimodal reasoning tasks. Our findings pave the way for more data-efficient and scalable multimodal models, demonstrating that structured knowledge transfer from pretrained vision models can serve as a viable alternative to large-scale multimodal pretraining from scratch.

{\small
\bibliographystyle{ieeenat_fullname}
\bibliography{11_references}
}

\ifarxiv \clearpage \appendix \section{Appendix Section}
\label{sec:appendix_section}

\subsection{Ablation on Combination Design}
\label{sec:appendix_comb}

\begin{table*}[h!]
\centering
% \footnotesize
\small
\begin{tabular}{cccccc}
\toprule
\textbf{Stride Size} & \textbf{Align Method} & \textbf{SQA$^\text{I}$} & \textbf{TQA} & \textbf{GQA} & \textbf{MME} \\ 
\midrule
4 \& 2 & Concat & 62.8 & 41.0 & 50.8 & 1278.1 \\
2 \& 4 & Concat & 63.2 & 41.2 & 51.2 & 1290.4 \\
2 \& 4 & AvgPool & 63.1 & 40.9 & 51.0 & 1262.7 \\
3 \& 4 & Concat & 63.7 & 41.4 & 51.7 &  1333.4 \\
\bottomrule
 \end{tabular}
    \caption{Ablation study of BREEN on the combination design for learnable queries with different granularity.}
    \label{table:ablation-combine}
\end{table*}

To optimize the performance when inclusion of multi-grained learnable queries, we ablate on different design choice for the combination as shown in Table~\ref{table:ablation-combine}. From the experiment results, we observe that both the order of the granularity and the combination method matters. From the first row and the second row, there is a performance drop when the finer-grained learnable queries are placed nearer the instruction text and farer from the images, which is intuitive as images are much more fine-grain than high-level text in terms of semantics. Therefore, placing finer-grained query nearer the images can have a better transition from low-level to high-level understanding.

We also try not concatenating the learnable queries of different granularity together, but align them from the same fine-grained sequence before and after average pooling according to the unflattened sequence, which is shown in the third row of Table~\ref{table:ablation-combine}. The performance also draws a little compared to concatenate different learnable queries together.

And finally, we choose concatenate learnable queries with stride size 3 and 4 together, which achieves highest score according to the experiments.

\subsection{Effect of Learnable Queries with Different Granularity}
\label{sec:appendix_gran}

\begin{table*}[ht]
\centering
% \footnotesize
\small
\begin{tabular}{cccccccccccc}
\toprule
\textbf{Stride/Patch Size} & \textbf{existence} & \textbf{count} & \textbf{position} & \textbf{color} & \textbf{posters} & \textbf{celebrity} & \textbf{scene}& \textbf{landmark} & \textbf{artwork} & \textbf{ocr} & \textbf{Total} \\ 
\midrule
2 ($12 \times 12$) & 173.3 & 88.3 & 78.3 & 101.6 & 56.5 & 47.1 & 139.0 & 82.5 & 85.8 & 67.5 & 919.9 \\

3 ($8 \times 8$) & 175.0 & 88.3 & 101.7 & 111.7 & 62.9 & 62.6 & 138.3 & 104.3 & 86.8 & 57.5 & 989.0 \\

4 ($6 \times 6$) & 175.0 & 90.0 & 98.3 & 145.0 & 68.4 & 47.4 & 134.3 & 85.3 & 90.8 & 57.5 & 991.8 \\

\bottomrule
 \end{tabular}
    \caption{Sub-task performance on MME-P.}
    \label{table:ablation-mmep}
\end{table*}

\begin{table*}[ht]
\centering
% \footnotesize
\small
\begin{tabular}{cccccc}
\toprule
\textbf{Stride/Patch Size} & \textbf{common sense reasoning} & \textbf{numerical 
calculation} & \textbf{text translation} & \textbf{code reasoning} & \textbf{Total} \\ 
\midrule
2 ($12 \times 12$) & 70.0 & 50.0 & 95.0 & 87.5 & 302.5 \\
 
3 ($8 \times 8$) & 72.1 & 42.5 & 110.0 & 80.0 & 304.64 \\ 

4 ($6 \times 6$) & 83.6 & 47.5 & 72.5 & 55.0 & 258.57 \\

\bottomrule
 \end{tabular}
    \caption{Sub-task performance on MME-C.}
    \label{table:ablation-mmec}
\end{table*}

We demonstrate more results on the inclusion of learnable queries with different granualarity on different sub-tasks of MME~\cite{Datasets:MME} in Table~\ref{table:ablation-mmep} and Table~\ref{table:ablation-mmec}. According to the results, we can find that finer-grained learnable queries (with patch size 2) can benefit more on the tasks requiring fine-grained recognition like OCR, numerical calculation and code reasoning. While coarse learnable queries will have an advantage on high-level understanding tasks, like color, posters, or artwork recognition and commonsense reasoning.

\subsection{Visualization}

\subsubsection{Visualization Overall Results with Radar Chart}
\label{sec:viz_rad}

\begin{figure}[t]
\centering
\small
\includegraphics[width=1\linewidth]
{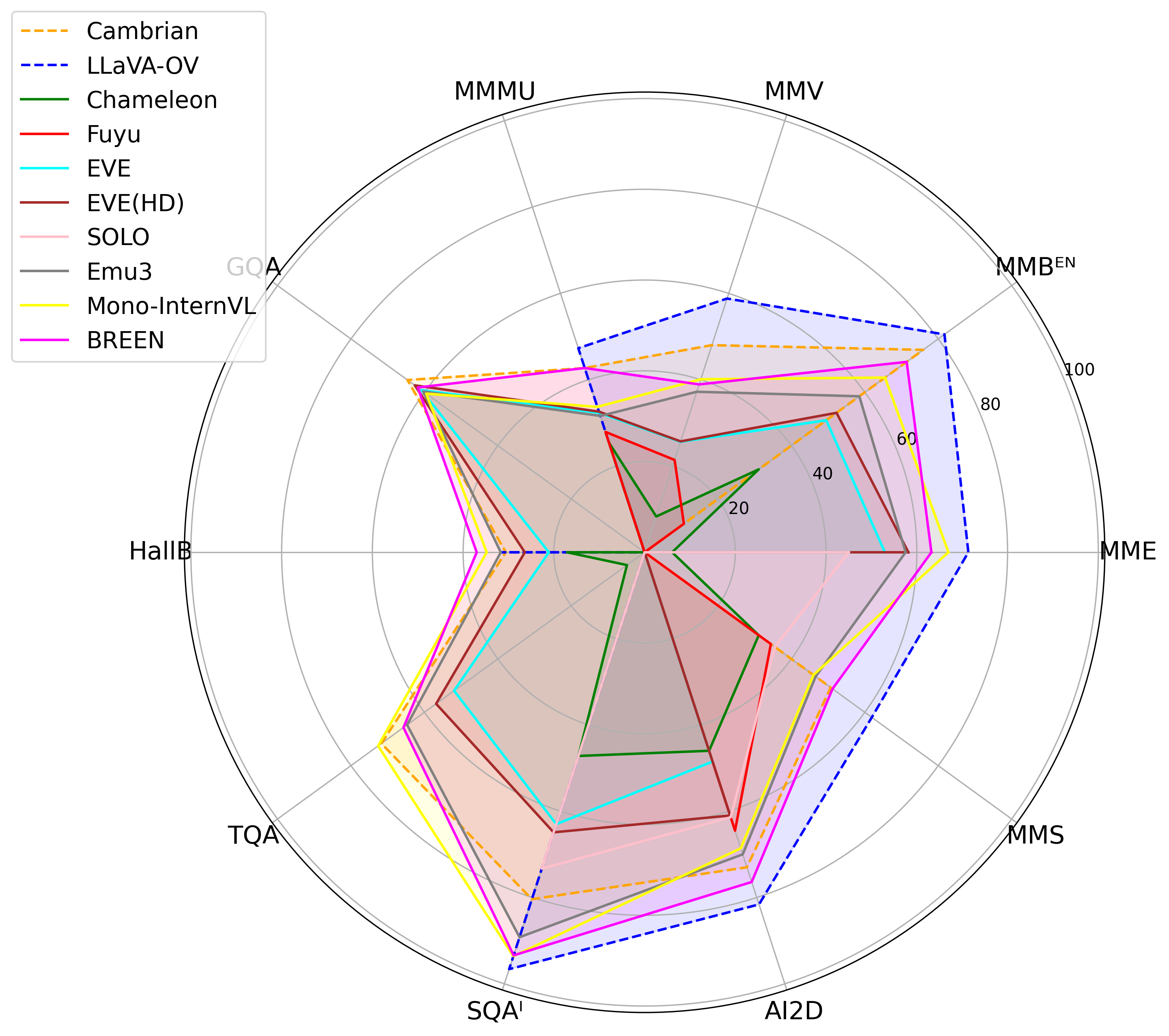}
  \caption{Comparison of the performance of BREEN with encoder-based and encoder-free models.}
\label{fig:model-perform}
\end{figure}

 We visualize the performance of encoder-free MLLMs and the selected top 2 encoder-based MLLMs in Figure~\ref{fig:model-perform}. It is obvious to see that BREEN performs comparable with a strong encoder-based baseline, which is Cambrian.

\subsubsection{Visualization of Attention Weights}
\label{sec:viz_attn}

\begin{figure*}[t]
    \centering 
    \includegraphics[width=0.98\linewidth,trim= 0 0 0 0,clip]
    {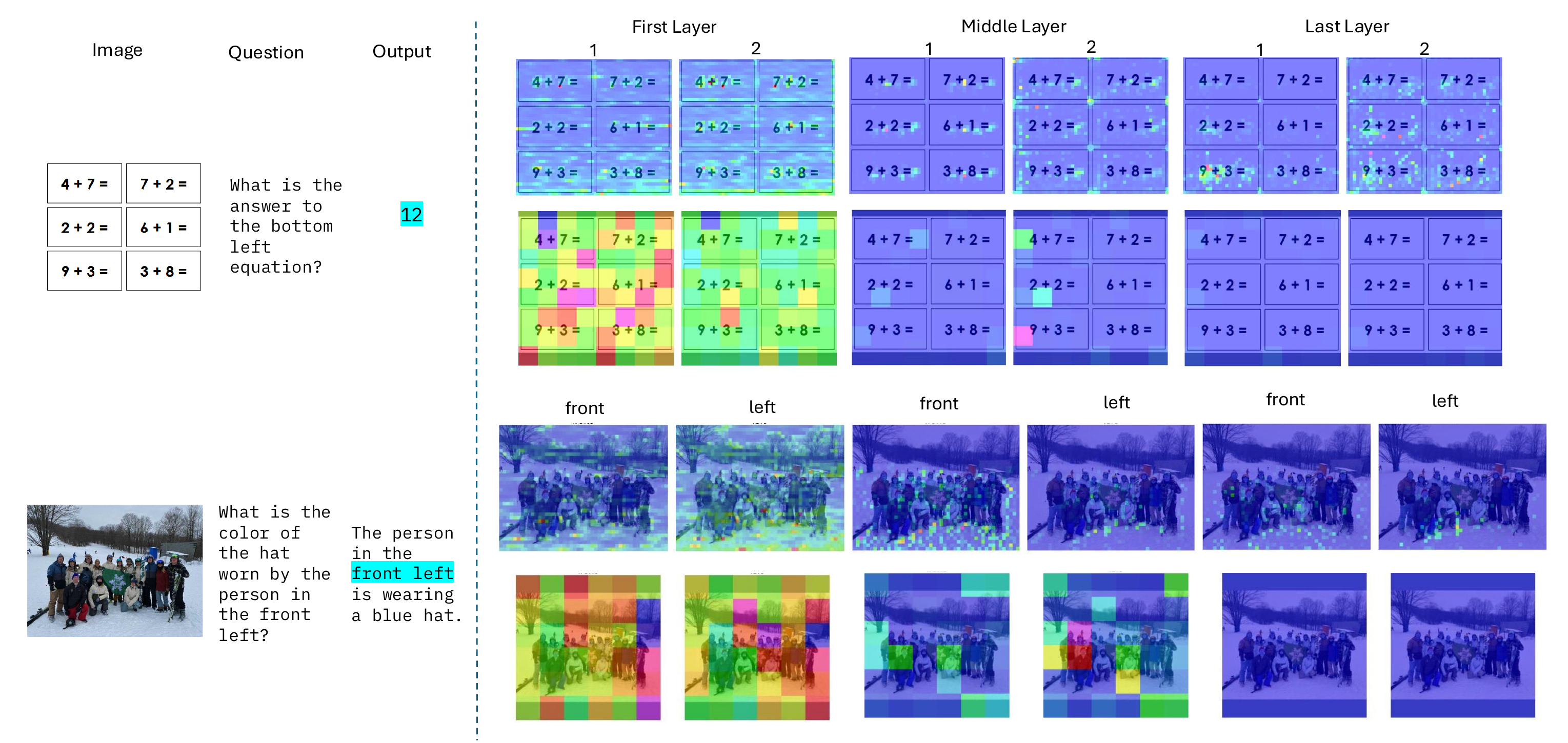} 
    \caption{The output attention visualization to image tokens and learnable query tokens in different layers.}
    \label{fig:attn_viz_app}
\end{figure*}

We visualize the attention weights from the critical output tokens to all the image tokens and the learnable query tokens in Figure~\ref{fig:attn_viz_app}. The first row of each example is the visualization for image tokens and the second row is for learnable queries. We can observe from the visualization that it is hard for attention weights to focus on the desired area in the first layer, they just distribute randomly with a higher average score. And we find that the attention to image tokens will refine to a specific area in the last layer, while still attending to a wider area in the middle layer. On the other hand, the attention scores to the learnable query can actively and correctly identify the important areas in the middle layer, but fade out in the last year. It is likely to be explained by the different granularity of the learnable query and image tokens.

\subsection{Limitation and Future Plan}
\label{sec:limiplan}

While BREEN presents a significant step toward data-efficient encoder-free multimodal learning, several limitations remain, providing avenues for future research.
\paragraph{Adaptive Query Selection for Task-Specific Needs}  
BREEN employs a dual-granularity learnable query mechanism, allowing the model to leverage both fine-grained and coarse-grained representations for different multimodal tasks. However, the current approach lacks an explicit mechanism for dynamically selecting or weighting queries based on the task context. Future work could explore adaptive query selection strategies, such as attention-based weighting or reinforcement learning-based query routing, to optimize query utilization based on task requirements.

\paragraph{Efficient Token Utilization and Sequence Compression}  
One limitation of BREEN is that the addition of learnable queries increases the token sequence length, leading to higher computational costs during training and inference. Although learnable queries improve vision-language alignment, their fixed length may introduce redundancy, especially for tasks with varying levels of visual detail. To address this, future work could explore mechanisms for adaptive query compression, where the model dynamically selects or merges learnable queries based on task requirements. Additionally, compressing image tokens alongside learnable queries could further enhance efficiency while retaining critical visual information.

\paragraph{Exploring Stronger Vision Encoder Teachers} BREEN distills visual knowledge from CLIP~\cite{Model:CLIP} to enhance alignment in an encoder-free setting. However, recent advancements in vision encoders, such as SigLIP and SigLIP2~\cite{Model:Siglip, Model:Siglip2}, have demonstrated superior visual representations with stronger semantic understanding. Future work could explore using these more powerful vision encoders as teachers to further improve the quality of learnable queries, potentially leading to better multimodal reasoning and more robust performance across diverse tasks.

 \fi

\end{document}